%% file: 0_main.tex
\documentclass[runningheads]{llncs}
\usepackage{graphicx}

\usepackage[table]{xcolor}
\usepackage{chngpage}
\usepackage{float}
\usepackage{etoolbox}
\usepackage{amsmath}
\usepackage{rotating}
\usepackage{graphicx}
\usepackage{svg}

\apptocmd{\sloppy}{\hbadness 10000\relax}{}{}

\begin{document}

\title{FedPID: An Aggregation Method for Federated Learning}

\input{1_authors.tex}

\maketitle
\setcounter{footnote}{0}
\begin{abstract}

This paper presents FedPID, our submission to the Federated Tumor Segmentation Challenge 2024 (FETS24). Inspired by FedCostWAvg and FedPIDAvg, our winning contributions to FETS21 and FETS2022, we propose an improved aggregation strategy for federated and collaborative learning. FedCostWAvg is a method that averages results by considering both the number of training samples in each group and how much the cost function decreased in the last round of training. This is similar to how the derivative part of a PID controller works. In FedPIDAvg, we also included the integral part that was missing. Another challenge we faced were vastly differing dataset sizes at each center. We solved this by assuming the sizes follow a Poisson distribution and adjusting the training iterations for each center accordingly. Essentially, this part of the method controls that outliers that require too much training time are less frequently used. Based on these contributions we now adapted FedPIDAvg by changing how the integral part is computed. Instead of integrating the loss function we measure the global drop in cost since the first round.


\vspace{0.6cm}
\end{abstract}
\newpage

\input{2_intro.tex}
\input{3_method.tex}

\input{4_results.tex}

\input{acknowledgment}

{ \bibliographystyle{splncs}
\bibliography{mybibliography}
}

\end{document}

%% file: 1_authors.tex

\author{Leon M\"achler\inst{3,4} \and
Gustav Grimberg\inst{3,4} \and
Ivan Ezhov\inst{1} \and
Manuel Nickel\inst{1}\and
Suprosanna Shit\inst{1} \and
David Naccache\inst{3} \and
Johannes C. Paetzold\inst{2}}

\authorrunning{M\"achler et al.}
\titlerunning{FedPIDAvg}

%
\institute{Department of Informatics, Technical University Munich, Germany\and 
Department of Computing, Imperial College London, London, UK\and
Département d'Informatique de l'ENS (DI ENS), École Normale Supérieure, PSL University, Paris, France \and
ezri ai labs\\
\email{leon.machler@ezri.ai}\\
}


%

%% file: 2_intro.tex
\section{Introduction}
    Federated learning provides an extremely promising framework for ensuring privacy and secure learning across different data locations \cite{yang2019federated}. It has numerous applications, ranging from power grids to medicine \cite{li2020review}. This approach is particularly important for medical images because patient information is highly sensitive, and the distribution of medical expertise as well as the prevalence of certain diseases varies significantly across different regions \cite{rieke2020future}. Additionally, medical imaging data is typically very large, making it expensive and impractical to frequently transfer from local clinics to a central server \cite{rieke2020future}. Ensuring the privacy and safety of patient data is even more crucial given the frequent illegal leaks of private medical records to the dark web \cite{meddatahack}. By keeping data local and only sharing updates, federated learning minimizes the risk of exposing sensitive information. This makes it an ideal solution for handling medical data securely and efficiently. Brain tumor segmentation is a particularly good candidate for study in this setting due to its complex and sensitive nature.

\begin{figure}[ht!]
    \label{fl}
    \centering
    \includegraphics[width=1\textwidth]{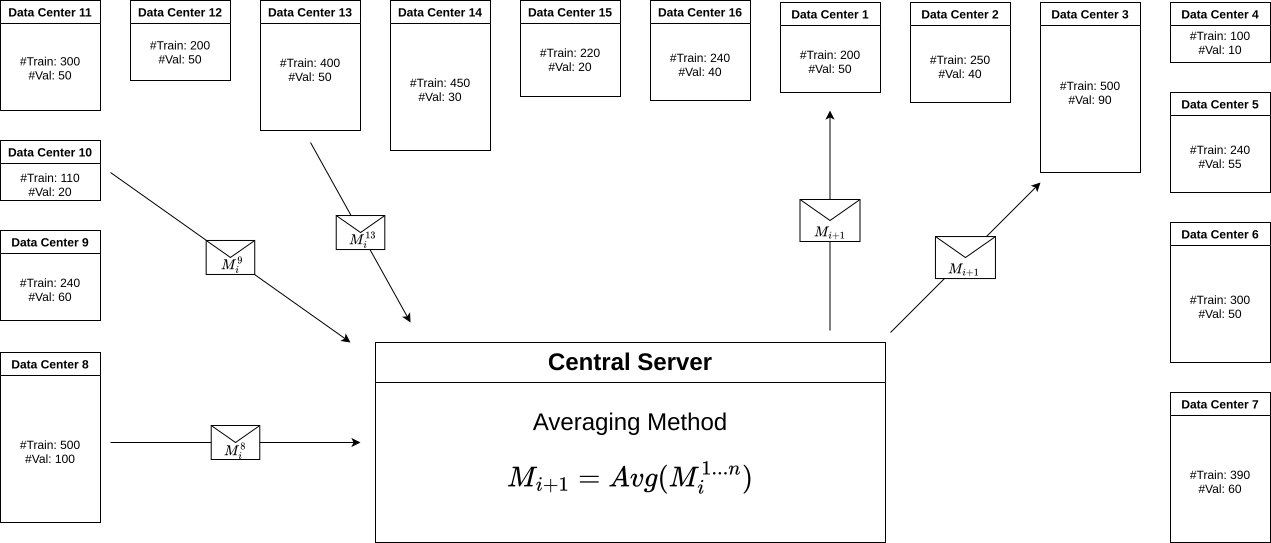}
    \caption{The schematic illustration shows the concept of federated learning. It depicts multiple data centers that form a federation. Each data center stores its training data of different sizes locally and trains the same model for a specific task, such as brain tumor segmentation in our case. During the aggregation step, the locally trained model weights are sent to a central server which then performs model aggregation and broadcasts the updated model back to the local centers. This process is repeated until the model converges or another stopping criterion is met.
    }
    \label{fig2}
\end{figure}

\subsection{FETS challenge}
The FETS challenge \cite{bakas2017advancing,pati2021federated,reina2021openfl,sheller2020federated,baid2021rsna} is a multi-year initiative aimed at addressing one of the main research question of federated learning: optimal aggregation of network weights from various data centers. In this paper, we address this issue by proposing a solution extending our previous submissions entailing PID controllers and classical statistics.

\section{Prior work}
This chapter delves into various established federated averaging techniques designed to enhance the efficiency and accuracy of brain tumor segmentation models across multiple data centers. We start with an overview of the traditional Federated Averaging (FedAvg) approach, which updates the global model by averaging local model weights, weighted by their respective training data sizes \cite{mcmahan2017communication}. We then introduce the Federated Cost Weighted Averaging (FedCostWAvg) method, a strategy that incorporates both data center sizes and cost function improvements, winning the FETS21 challenge \cite{machler2021fedcostwavg}. Lastly, we present Federated PID Weighted Averaging (FedPIDAvg), which integrates the integral term of a PID controller-inspired solution to further refine the model averaging process, outperforming all other submissions to the FETS22 challenge \cite{machler2022fedpidavg} in the process yet again. Each of the introduced methods aims to optimize model performance while maintaining data privacy and minimizing communication costs.

\subsection{Federated Averaging (FedAvg)}

The traditional federated averaging (FedAvg) approach \cite{mcmahan2017communication} employs an averaging strategy on the local model weights to update the global model, weighted by the training data set sizes of the local models. A model $M_{i+1}$ is  updated as 

\begin{equation}
    M_{i+1} = \frac{1}{S} \sum_{j=1}^n s_j M_{i}^j ,
\end{equation}\\

where $s_j$ denotes the number of samples that model $M^j$ was trained on in round $i$ and $S = \sum_j s_j$. The definition is adapted from \cite{machler2021fedcostwavg}.

\subsection{Federated Cost Weighted Averaging (FedCostWAvg)}
For the FeTS2021 challenge, we proposed a new weighting strategy which not only takes into account data center sizes but also the amount by which the cost function decreased during the last step \cite{machler2021fedcostwavg}. We termed this method FedCostWAvg, where the new model $M_{i+1}$ is calculated as

\begin{equation}
    M_{i+1} = \sum_{j=1}^n (\alpha \frac{s_j}{S} + (1 - \alpha) \frac{k_j}{K}) M_{i}^j
\end{equation}
with
\begin{equation}
     k_j = \frac{c(M_{i-1}^j)}{c(M_{i}^j)}, K = \sum_j k_j.
\end{equation}\\

Here, $c(M_i^j)$ denotes the cost of model $j$ at time-step $i$, computed from the local cost function \cite{machler2021fedcostwavg}.  The hyperparameter $\alpha \in [0,1]$ is chosen to balance the impact of the cost improvements and the data set sizes. In 2021, we won the challenge with an alpha value of \emph{$\alpha = 0.5$}. The FedCostWAvg weighting strategy adjusts for the training dataset size and also for local improvements in the last training round.

\subsection{Federated PID Weighted Averaging (FedPIDAvg)}
Introducing FedCostWAvg in \cite{machler2021fedcostwavg}, we recognized its similarity to a PID controller with the only missing element being the integral term. The subsequently proposed FedPIDAvg was novel in two ways: It introduced a PID-inspired integral term as well as a different way to calculate the differential term.
FedPIDAvg computes the new model $M_{i+1}$ as
\begin{equation}
\label{equ:fedpidavg}
    M_{i+1} = \sum_{j=1}^n (\alpha \frac{s_j}{S} + \beta \frac{k_j}{K} + \gamma \frac{m_j}{I} ) M_{i}^j
\end{equation}
where
\begin{equation}
     k_j = c(M_{i-1}^j) - c(M_{i}^j), K = \sum_j k_j
\end{equation}
states the reduction in cost between subsequent rounds and
\begin{equation}
     m_j = \sum_{l=0}^5 c(M_{i-l}) , I = \sum_j m_j
\end{equation}\\
integrates cost values over a history of rounds. All terms in Equation \ref{equ:fedpidavg} are scaled by a set of hyperparameters, $(\alpha, \beta, \gamma)$, which are normalized such that 
\begin{equation}
    \alpha + \beta + \gamma = 1
\end{equation}\\
always holds.

Note that in FedPIDAvg, we use the absolute difference between the last and the new cost rather than the ratio as previously proposed for FedCostWAvg. The FedPIDAvg strategy remains a weighted averaging approach, where the weights are calculated as weighted averages of three factors: the drop in cost from the last round, the sum of costs over the last several rounds, and the size of the training dataset. These factors are weighted by $\alpha$, $\beta$, and $\gamma$, respectively. The values of $\alpha$, $\beta$, and $\gamma$ need to be optimized based on the specific use case. In 2022, we chose $0.45$, $0.45$, and $0.1$; we did not have the resources to study the sensitivity to these parameter choices. This method won FETS2022.

%% file: 3_method.tex
\section{Methodology}

In the following chapter, we first introduce and formalize our novel averaging concept named \textit{FedPID}. We then briefly describe the neural network architecture for brain tumor segmentation that was given by the challenge organizers and finally discuss our strategy regarding when to train and aggregate from which specific centers, depending on their training samples modeled using a simple Poisson distribution. 

\subsection{Improved Federated PID Weighted Averaging (FedPID)}

This year's method, FedPID, is a refinement of FedPIDAvg.  We calculate the new model $M_{i+1}$ in the following manner: 

    \begin{equation}
    M_{i+1} = \sum_{j=1}^n (\alpha \frac{s_j}{S} + \beta \frac{k_j}{K} + \gamma \frac{m_j}{I} ) M_{i}^j.
\end{equation}
with:
\begin{equation}
     k_j = c(M_{i-1}^j) - c(M_{i}^j), K = \sum_j k_j.
\end{equation}
and:
\begin{equation}
     m_j = \frac{c(M_{1})}{c(M_{i})}, I = \sum_j m_j.
\end{equation}\\
\begin{equation}
    \alpha + \beta + \gamma = 1
\end{equation}\\

The notable difference is the way the I term is computed. We take the ratio of the cost of the second round divided by that of the current round. This measures similar to the D term the progress but on a global scale. We chose the second round to avoid outliers at the start.


\subsection{U-Net for Brain Tumor Segmentation}
A 3D-Unet was provided as segmentation network by the challenge organizers, a highly successful neural network architecture in medical image analysis \cite{ronneberger2015u}. Figure \ref{fig2} shows a schematic illustration of its architecture, modifications to which were disallowed for the entirety of the challenge.
U-Nets are considered state-of-the-art for a wide range of applications, including brain tumor segmentation \cite{menze2014multimodal,bakas2017advancing}, vessel segmentation \cite{vessap,cldice}, and many more.

\begin{figure}[t!]
    \centering
    \includegraphics[width=\textwidth]{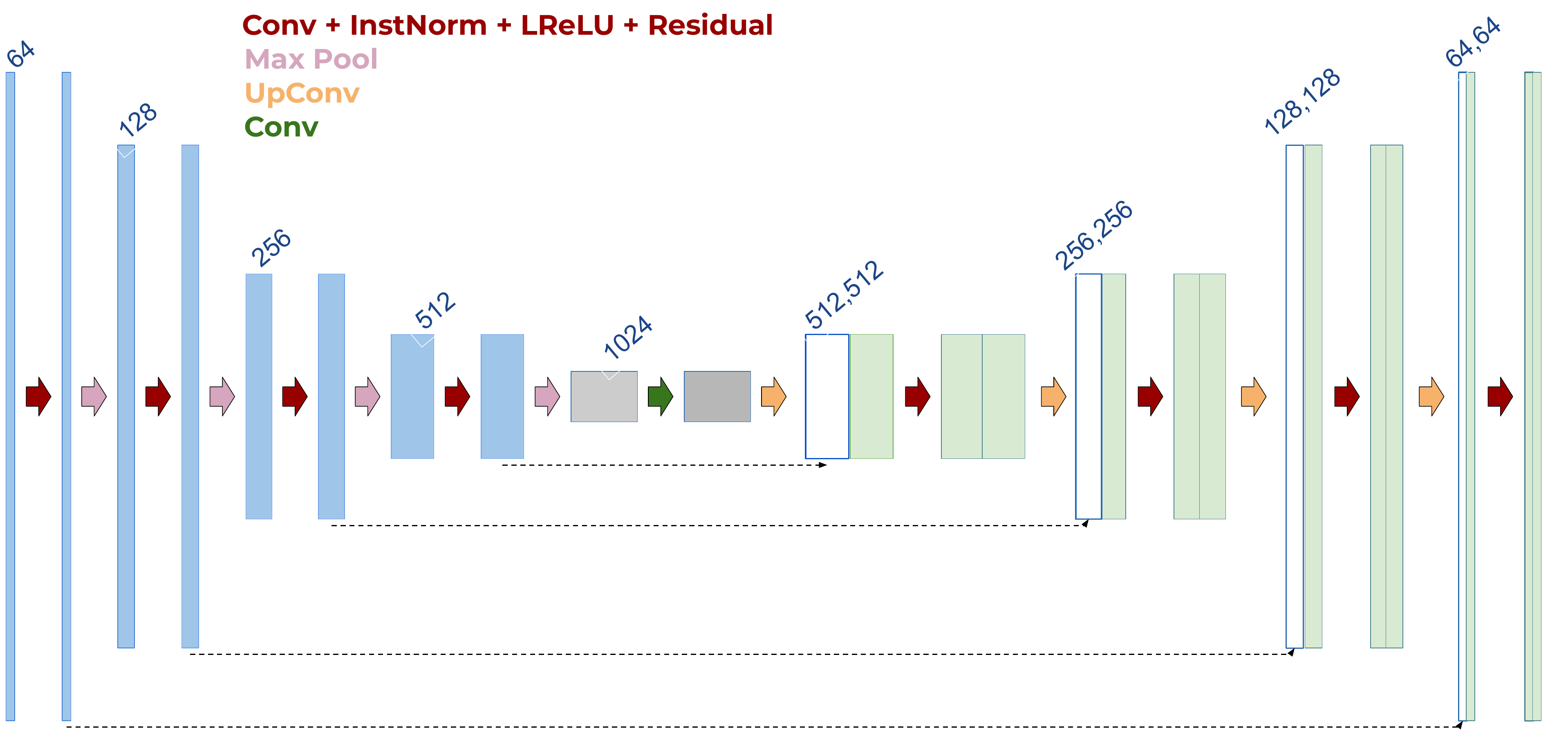}
    \caption{The common 3D U-net architecture, which is used in many medical imaging tasks. It was provided as such by the FETS challenge; modifications were not allowed \cite{ronneberger2015u,pati2021federated}.}
    \label{unet}
\end{figure}

\subsection{Poisson-distribution modeling of the data samples per center}
In order to optimize training speed over several federated rounds, it was possible to only select a subset of data centers for each one. In our initial submission \cite{machler2021fedcostwavg}, we simply selected all centers every time. The year after, we added a novel way to select participating data centers at each federated round. To do so, we resorted to classical statistics means, namely, under the assumption that the dataset sizes follow a Poisson distribution:
\begin{equation}
\begin{aligned}
p(x;\lambda) = \frac{e^{-\lambda}\lambda^{x}} {x!}  \\ \\ \mbox{    with     } 
x = 0, 1, 2, \cdots
\end{aligned}
\end{equation} \\
we made the natural choice of dropping outliers in most rounds, where outliers were defined as having $x>2\lambda$. For this year's submission, we chose $x>\lambda$ but also made sure to always use at least 50\% of all collaborators.



%% file: 4_results.tex
\section{Results}
In this section, we are presenting the results of our methods on the FeTS challenges 2021, 2022 and 2024. \\

"FeTS borrows its data from the BraTS Continuous Evaluation, but additionally providing a data partitioning according to the acquisition origin for the training data. Ample multi-institutional, routine clinically-acquired, pre-operative baseline, multi-parametric Magnetic Resonance Imaging (mpMRI) scans of radiographically appearing glioblastoma (GBM) are provided as the training and validation data for the FeTS 2022 challenge. Specifically, the datasets used in the FeTS 2022 challenge are the subset of GBM cases from the BraTS Continuous Evaluation. Ground truth reference annotations are created and approved by expert board-certified neuroradiologists for every subject included in the training, validation, and testing datasets to quantitatively evaluate the performance of the participating algorithms." \cite{fetswebsite}

In Tables \ref{tab:f21.1} and \ref{tab:f21.2}, we provide the results of FedCostWAvg in the 2021 challenge. Tables \ref{tab:f22.1} and \ref{tab:f22.2} show the results of FedPIDAvg in the FETS 2022 challenge. Amongst all submitted methods in 2021 and 2022, our methods performed best and won the respective challenges. Note that the data varies from year to year; this influences the available rounds of federated training.

\begin{table}[H]
\begin{adjustwidth}{0in}{0in}
    \centering
    \setlength\tabcolsep{3pt} 
    \begin{tabular}{|l|c|c|c|c|c|c|c|c|c|c|c|c|c|c|}
        \hline
        
        Label & Dice WT & Dice ET & Dice TC & Sens. WT & Sens. ET & Sens. TC   \\ \hline
        Mean & 0,8248 & 0,7476 & 0,7932 & 0,8957 & 0,8246 & 0,8269  \\ \hline 
        StdDev & 0,1849 & 0,2444 & 0,2643 & 0,1738 & 0,2598 & 0,2721  \\ \hline
        Median & 0,8936 & 0,8259 & 0,9014 & 0,948 & 0,9258 & 0,9422  \\ \hline
        25th quantile & 0,8116 & 0,7086 & 0,8046 & 0,9027 & 0,7975 & 0,8258 \\ \hline
        75th quantile & 0,9222 & 0,8909 & 0,942 & 0,9787 & 0,9772 & 0,9785  \\ \hline
    \end{tabular}
    \vspace{2pt}
    \caption{\label{tab:f21.1} Final performance of FedCostWAvg in the 2021 FETS Challenge, DICE and Sensitivity}
\end{adjustwidth}
\end{table}

\begin{table}[H]
\begin{adjustwidth}{0in}{0in}
    \centering
    \setlength\tabcolsep{3pt} 
    \begin{tabular}{|l|c|c|c|c|c|c|c|c|c|c|c|c|c|c|}
        \hline
        Label & Spec WT & Spec ET & Spec TC & H95 WT & H95 ET & H95 TC & Comm. Cost  \\ \hline
        Mean & 0,9981 & 0,9994 & 0,9994 & 11,618 & 27,2745 & 28,4825 & 0,723 \\ \hline
        StdDev & 0,0024 & 0,0011 & 0,0014 & 31,758 & 88,566 & 88,2921 & 0,723 \\ \hline
        Median & 0,9986 & 0,9996 & 0,9998 & 5 & 2,2361 & 3,0811 & 0,723 \\ \hline
        25th quantile & 0,9977 & 0,9993 & 0,9995 & 2,8284 & 1,4142 & 1,7856 & 0,723 \\ \hline
       75th quantile & 0,9994 & 0,9999 & 0,9999 & 8,6023 & 3,5628 & 7,0533 & 0,723 \\ \hline

   \end{tabular}
   \vspace{2pt}
   \caption{\label{tab:f21.2} Final performance of FedCostWAvg in the 2021 FETS Challenge, Specificity, Hausdorff95 Distance and Communication Cost}
\end{adjustwidth}
\end{table}

\begin{table}[H]
\begin{adjustwidth}{0in}{0in}
    \centering
    \setlength\tabcolsep{3pt} 
    \begin{tabular}{|l|c|c|c|c|c|c|c|c|c|c|c|c|c|c|}
        \hline
        Label & Dice WT & Dice ET                & Dice TC         & Sens WT         & Sens ET         & Sens TC  \\ \hline 
        Mean & 0,76773526 & 0,741627265          & 0,769244434     & 0,749757737     & 0,770377324     & 0,765940502  \\ \hline  
        StdDev & 0,183035406 & 0,266310234       & 0,284212379     & 0,208271565     & 0,280923214     & 0,297081407 \\ \hline  
        Median & 0,826114563 & 0,848784494       & 0,896213442     & 0,819457864     & 0,886857246     & 0,893165349  \\ \hline   
        25quant & 0,700757354 & 0,700955694      & 0,739356651     & 0,637996476     & 0,728202272     & 0,73357786 \\ \hline   
        75quant & 0,897816734 & 0,910451814      & 0,943718628     & 0,905620122     & 0,956570051     & 0,964129538 \\ \hline  
        
    \end{tabular}
    \vspace{2pt}
    \caption{\label{tab:f22.1} Final performance of FedPIDAvg in the 2022 FETS Challenge, DICE and Sensitivity}
\end{adjustwidth}
\end{table}

\begin{table}[H]
\begin{adjustwidth}{0in}{0in}
    \centering
    \setlength\tabcolsep{3pt} 
    \begin{tabular}{|l|c|c|c|c|c|c|c|c|c|c|c|c|c|c|}
        \hline
        Label   & Spec WT    & Spec ET   & Spec TC           & H95 WT    & H95 ET          & H95 TC & Comm. Cost  \\ \hline
        Mean  & 0,9989230   & 0,9995742  & 0,999692          & 24,367549 & 32,796706    & 32,466108 & 0,300 \\ \hline
        StdDev  & 0,0016332  & 0,0007856   & 0,0006998       & 32,007897 & 89,31835     & 85,440174 & 0,300 \\ \hline
        Median   & 0,9994479   & 0,9997990  & 0,999868       & 11,57583  & 2,4494897     & 4,5825756 & 0,300 \\ \hline
        25th quant   & 0,998690  & 0,9995254  & 0,999695     & 5,4081799 & 1,4142135    & 2,2360679 & 0,300 \\ \hline
       75th quant  & 0,9998540   & 0,9999292   & 0,9999584   & 36,454261 & 10,805072    & 12,359194 & 0,300 \\ \hline

   \end{tabular}
   \vspace{2pt}
   \caption{\label{tab:f22.2} Final performance of FedPIDAvg in the 2022 FETS Challenge, Specificity, Hausdorff95 Distance and Communication Cost}
\end{adjustwidth}
\end{table}

    


    


    


Finally,  we are providing the results of the 2024 challenge as soon as they are published on the challenge website.

\section{Conclusion}

This paper summarizes our contribution to the Federated Tumor Segmentation Challenge 2024 and discusses our team's previous submissions to the challenges in 2021 and 2022. This year, we submitted a method that is a refined version of the 2022 submission by changing the I term to track global progress. 
As in our 2022 submission, the client selection models data center sizes using a Poisson distribution, which essentially excludes the outliers.



%% file: acknowledgment.tex
\section*{\large{Acknowledgements}}
\noindent We appreciate the valuable input from the senior scientists, David Naccache, Adrian Dalca, Georgios Kaissis, Daniel Rückert and Bjoern Menze. Moreover, we want to express our appreciation to the organizers of the Federated Tumor Segmentation Challenge. Leon M\"achler and Gustav Grimebrg are supported by the École normale supérieure in Paris and run ezri ai labs. Johannes C. Paetzold is supported by Imperial College London. Suprosanna Shit, Manuel Nickel and Ivan Ezhov are supported by the Technical University of Munich – Institute for Advanced Study, funded by the German Excellence Initiative. Ivan Ezhov is also supported by the International Graduate School of Science and Engineering (IGSSE). 